\documentclass{article} 
\usepackage{booktabs}
\usepackage{multirow}
\usepackage{array}
\usepackage{iclr2027_conference,times}

\usepackage{amsmath,amsfonts,bm}

\def\eqref#1{equation~\ref{#1}}

\def\1{\bm{1}}

\DeclareMathAlphabet{\mathsfit}{\encodingdefault}{\sfdefault}{m}{sl}
\SetMathAlphabet{\mathsfit}{bold}{\encodingdefault}{\sfdefault}{bx}{n}

\usepackage{hyperref}
\usepackage{url}
\usepackage{xspace}
\usepackage{booktabs}
\usepackage{graphicx}
\title{DLite: History-Free Speculative Decoding for Agentic AI}

\author{
\normalfont
\textbf{Jiaming Xu}\textsuperscript{1,2,*}\quad
\textbf{Hanzhen Wang}\textsuperscript{1,}\thanks{The first two authors contributed equally to this work}\quad
\textbf{Tianlang Zhao}\textsuperscript{1}\quad
\textbf{Jiewen Xiao}\textsuperscript{1}\quad
\textbf{Chengze Yuan}\textsuperscript{1} \\
\normalfont
\textbf{Jiayi Pan}\textsuperscript{1}\quad
\textbf{Yongkang Zhou}\textsuperscript{1,2}\quad
\textbf{Jiancai Ye}\textsuperscript{1}\quad
\textbf{Yu Wang}\textsuperscript{3}\quad
\textbf{Guohao Dai}\textsuperscript{1,2,4,}
\thanks{Correspondence to Guohao Dai (Email Address: daiguohao@sjtu.edu.cn)}
\\[0.5em]
\normalfont\textsuperscript{1}Shanghai Jiao Tong University \\
\normalfont\textsuperscript{2}Shanghai Innovation Institute \\
\normalfont\textsuperscript{3}Tsinghua University \\
\normalfont\textsuperscript{4}Infinigence-AI
}

\newcommand{\we}{\emph{DLite}\xspace}

\iclrfinalcopy 

\begin{document}

\begingroup
\iclrfinalcopy
\maketitle
\endgroup
\fancyhead{}

\begin{abstract}

Speculative decoding has become a mainstream technique for accelerating large language model (LLM) inference by using a lightweight draft model to generate candidate tokens that are subsequently verified by the target model. However, agentic workloads introduce long and continuously evolving contexts and frequent multi-turn interactions. Under such workloads, we observe that existing draft models suffer from up to an 80\% degradation in acceptance length, while maintaining their historical KV cache introduces substantial system overhead. In this paper, we identify the growing historical context as a common source of both problems and introduce \we, a history-free speculative decoding framework for agentic AI. \we directly reuses contextual hidden states from the target model and performs local speculative prediction without explicitly maintaining the historical context, transforming an increasingly complex modeling problem into local simple prediction problem. A lightweight sequential head further restores intra-block causal dependencies with low overhead. To effectively train this heterogeneous history-free architecture, we introduce a tailored two-stage training strategy. Experiments on general-purpose, long-context and agentic workloads show that \we achieves $2 \sim 6$ acceptance length and outperforms the state-of-the-art speculative decoding method, DSpark, up to $70\%$.

\end{abstract}

\begin{center}
\urlstyle{same}
\vspace{-10pt}
\begin{tabular}{@{}c@{\hspace{0.8em}}l@{}}
\raisebox{-0.25em}{\includegraphics[height=1.5em]{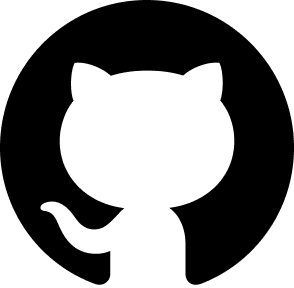}} &
\url{https://github.com/SJTU-DAI-Sys/DLite} \\[0.3em]
\raisebox{-0.25em}{\includegraphics[height=1.5em]{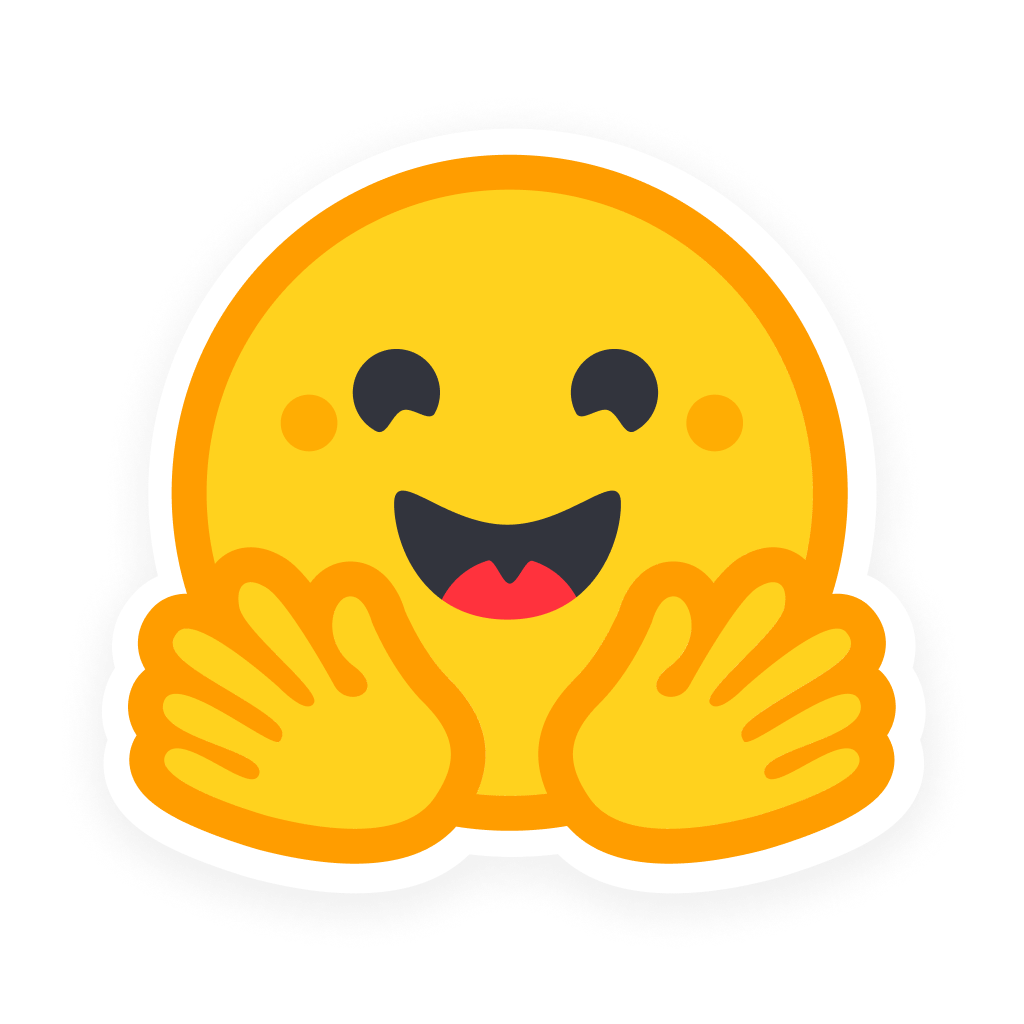}} &
\url{https://huggingface.co/collections/SJTU-DAI-Sys/dlite}
\end{tabular}
\end{center}

\section{Introduction}

Large language models (LLMs) have become the foundation of modern AI agents in applications such as coding, web search, and medicine. As agent execution increasingly relies on frequent LLM invocations, inference efficiency has become a critical factor in both system throughput and user experience. Consequently, extensive research~\cite{zhou2024survey,li2024large} has explored various techniques to achieve extreme optimization on LLM inference, encompassing algorithm optimization~\cite{chen2026dflash,cheng2026dspark}, system enhancements~\cite{dao2022flashattention,hong2024flashdecoding++}, and hardware advancements~\cite{moon2024latency,prabhakar2024sambanova}.


Speculative decoding has emerged as a mainstream inference acceleration technique, which has been widely adopted by prominent LLMs (\textit{e.g.}, DeepSeek-V4~\cite{xu2026deepseek}) and deployed in real-world serving systems(\textit{e.g.}, SGLang~\cite{zheng2024sglang} and vLLM~\cite{kwon2023efficient}). A lightweight draft model first generates multiple candidate tokens, which are then verified by the target model in a single forward pass. By accepting multiple tokens at once, speculative decoding can substantially improve inference efficiency without compromising accuracy.

\begin{figure}[!t]
    \centering
    \includegraphics[width=0.98\textwidth]{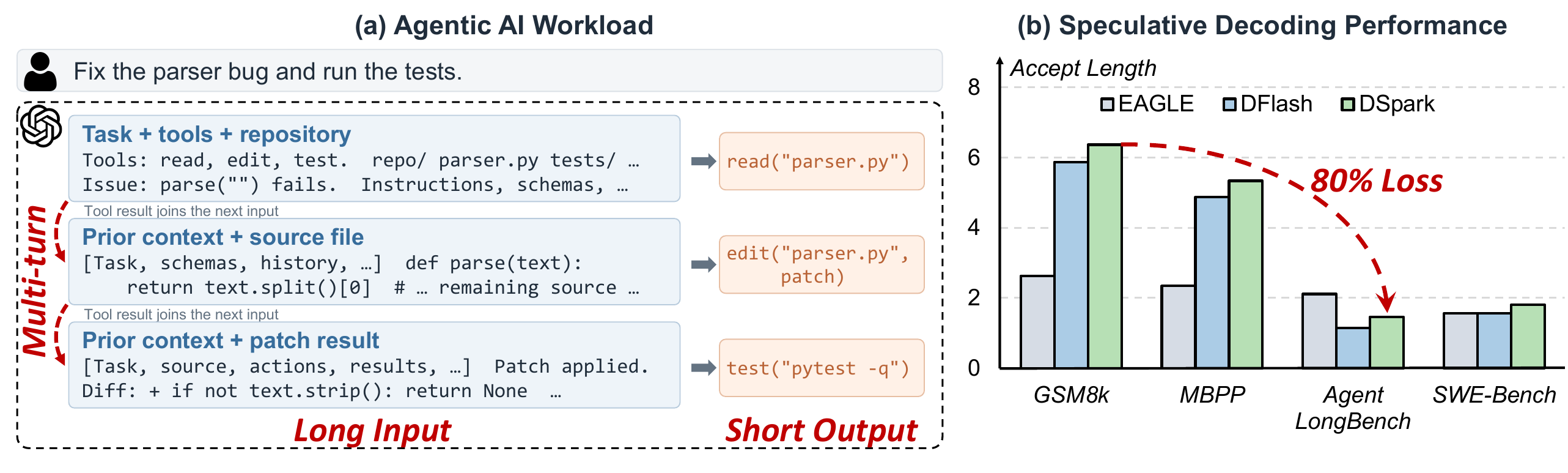}
    \caption{(a) Agentic workloads characteristics. (b) Speculative decoding performance loss.}
    \vspace{-10pt}
    \label{fig:agent}
\end{figure}

However, agentic workloads differ significantly from conventional conversational inference. As illustrated in Figure~\ref{fig:agent}(a), agent execution typically involves long inputs, short outputs, frequent multi-turn interactions, and continuously growing contexts. Under such workloads, we observe two major limitations of existing speculative decoding methods. \textbf{(1) From the perspective of algorithmic performance},  Figure~\ref{fig:agent}(b) shows that the methods that achieve high acceptance lengths on conventional benchmarks can drop to only \(1\sim2\) accepted tokens on agentic workloads, corresponding to a degradation of more than \(80\%\). Lightweight draft models increasingly struggle to independently model the long historical context required to match the predictive distribution of the target model. \textbf{(2) From the perspective of system efficiency}, maintaining the draft model's historical states introduces non-negligible system overhead. Existing speculative drafters must continuously update their KV cache with newly accepted tokens and maintain historical states for subsequent generation and prefix reuse. Under frequent multi-turn agent execution, these operations can account for \(10\%\sim20\%\) of end-to-end inference latency.

These two limitations share a common source: the draft model is required to independently encode and maintain an increasingly long history, though the target model has processed the same context. More importantly, this growing history continuously increases the explicit context length and positional range faced by the lightweight drafter, making the prediction problem increasingly difficult. We therefore adopt a different design principle: \textbf{rather than reconstructing the full history inside the draft model, the drafter directly reuses contextual representations produced by the target model and performs only local speculative prediction around the current decoding position.} 

Based on this insight, we propose \we, a history-free speculative decoding framework for agentic AI. \we employs a history-free parallel backbone that directly consumes target hidden states and performs local prediction around the current decoding position without maintaining historical draft KV cache. A lightweight sequential head restores the missing intra-block causal dependency directly in the hidden space, avoiding expensive intermediate vocabulary projections. Finally, we introduce a two-stage training strategy that first bridges the structural mismatch using an isomorphic history-aware teacher and then directly aligns \we with the target model. 

We train \we for Qwen3-4B, Qwen3-8B models and evaluate it on both general-purpose, long-context and agentic workloads. \we achieves \(2\sim6\) acceptance length and improves over DSpark by up to \(70\%\), completely eliminating the draft model's historical KV-cache maintenance overhead.

\section{Background}


\subsection{Speculative Decoding}
\begin{figure}[!t]
    \centering
    \includegraphics[width=0.98\textwidth]{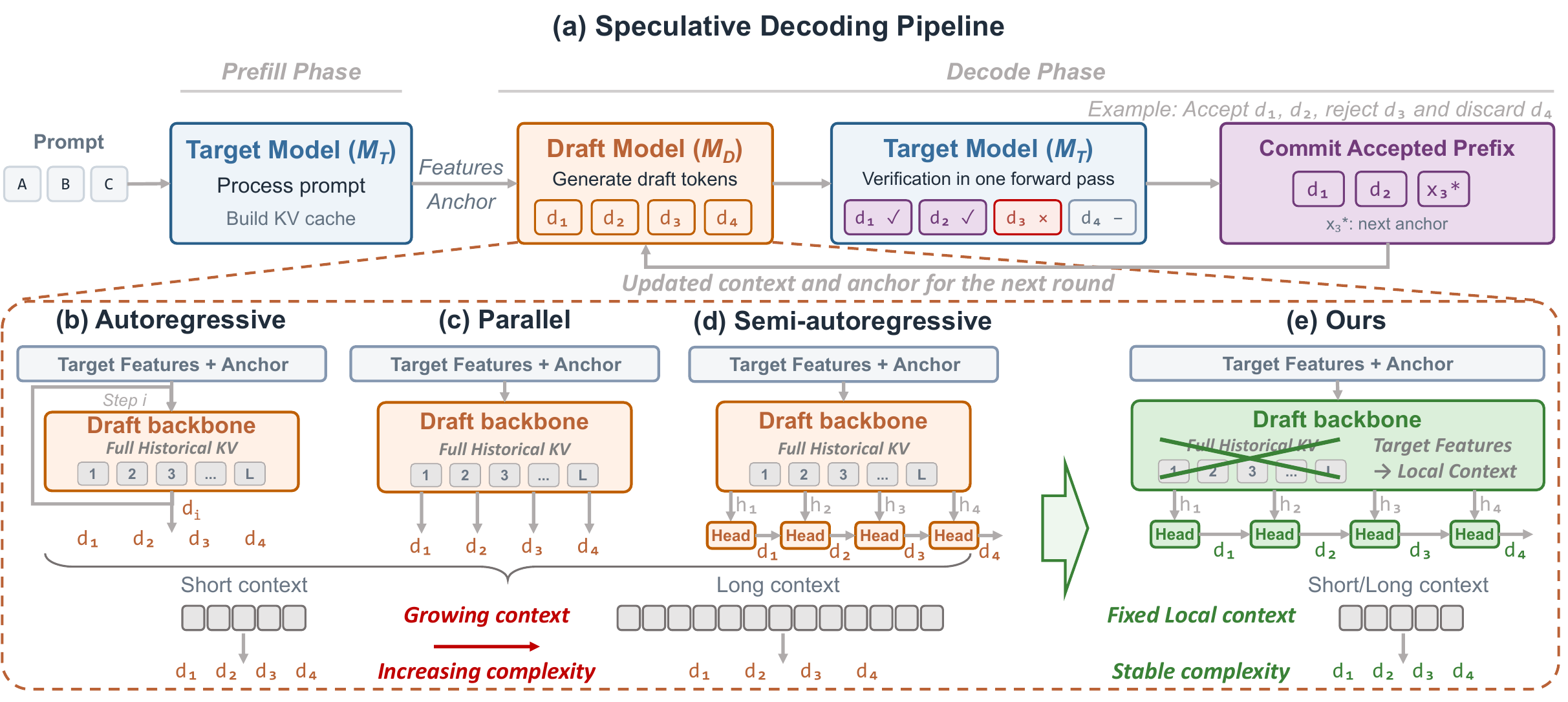}
    \caption{Speculative decoding pipeline and comparison of existing draft architectures with \we.}
    \label{fig:background}
    \vspace{-15pt}
\end{figure}

Speculative decoding accelerates LLM inference through a draft-and-verify pipeline shown in Figure~\ref{fig:background}(a). Given the current context, a lightweight draft model \(M_D\) generates \(N\) candidate tokens, which are subsequently verified in parallel by the target model \(M_T\). If \(\tau\) tokens are produced per verification round, the practical speedup~\cite{nvidiaspec} can be approximately expressed as

\begin{equation}
    \mathrm{Speedup} = \frac{\mathrm{Time}_{M_T}(1) \times \tau}{\mathrm{Time}_{M_T}(N)+\mathrm{Time}_{M_D}(N)}
\end{equation}

Therefore, speculative decoding performance mainly depends on two factors: acceptance length \(\tau\) and draft-model latency \(\mathrm{Time}_{M_D}(N)\).

Existing draft models mainly follow three architectural paradigms, summarized in Figure~\ref{fig:background}(b)(c)(d). Autoregressive drafters, represented by EAGLE~\cite{li2026eagle3}, explicitly preserve causal dependencies among speculative tokens but require sequential draft steps. Parallel drafters, such as DFlash~\cite{chen2026dflash}, predict all speculative positions in one forward pass, substantially reducing drafting latency but removing intra-block causal dependency. Semi-autoregressive drafters, including DSpark~\cite{cheng2026dspark} and Domino~\cite{huang2026domino}, combine a parallel backbone with a lightweight sequential module to balance drafting latency and prediction quality. Despite their different structures, these designs share one property: the draft model still explicitly depends on the historical context, whose length grows with decoding. In contrast, Figure~\ref{fig:background}(e) shows that \we removes this growing historical dependency by directly reusing contextual features produced by the target model and performs local prediction.

\subsection{Model Distillation}

Training a speculative draft model can be viewed as knowledge distillation from the target model. Different architectures adopt different supervision objectives. EAGLE performs distribution-level distillation using soft-target cross-entropy (CE), which is equivalent to minimizing forward Kullback-Leibler divergence up to a constant. DFlash primarily adopts token-level CE, while DSpark combines CE with total variation (TV) distance to improve distribution overlap between the draft and target models. Domino applies token-level supervision to both parallel and causally corrected predictions through a training curriculum. These differences are particularly relevant to \we because its history-free prediction structure differs substantially from the autoregressive target model, making direct distribution-level distillation non-trivial.

\section{Motivation} \label{sec:motivation}

\subsection{Two Important Observations}

Agentic workloads expose two limitations of existing speculative decoding methods: degraded acceptance length and increasing historical-state maintenance overhead.


\textbf{Degraded Acceptance Length.} As illustrated in Figure~\ref{fig:agent}(b), existing speculative decoding methods achieve high acceptance lengths on general workloads such as GSM8K and MBPP, but degrade substantially on agentic workloads such as SWE-Bench and AgentLongBench. As the context grows, lightweight draft models increasingly struggle to capture the information required to match the target model's predictions, causing acceptance lengths to drop to only \(1\sim2\) tokens in some workloads.


\textbf{KV Cache Maintenance.} Existing speculative drafters must maintain historical KV states for subsequent generation. After each target-model verification, accepted tokens are committed to the draft model to update its KV cache, while prefix caching further requires historical draft states to be preserved across agent turns. Under frequent multi-turn workloads with continuously growing contexts, these operations become increasingly expensive. Using SGLang to evaluate the multi-token prediction module in DeepSeek-V4, we observe that the KV cache  updating and maintenance can account for \(10\%\sim20\%\) of end-to-end inference latency.


\subsection{Analysis and Findings}




Although the above problems appear at different levels, they share a common cause: the draft model is required to independently process and maintain the growing historical context shown in Figure~\ref{fig:background}. From the model perspective, a lightweight drafter has only a fraction of the target model's capacity, making it increasingly difficult to reproduce the target model's long-context representation as history grows. From the system perspective, preserving the same history requires maintaining and repeatedly updating an increasingly large draft KV cache.

The key inefficiency is therefore that the draft model repeatedly reconstructs historical information that has already been processed by the target model. We instead argue that the drafter should directly reuse the target model's contextual representations and dedicate its limited capacity to modeling the information required for local speculative prediction. 

\section{\we Framework}

Motivated by the observations in Section~\ref{sec:motivation}, we design \we to decouple historical-context modeling from speculative generation. As illustrated in Figure~\ref{fig:background}(e), instead of requiring a lightweight draft model to repeatedly process and maintain an ever-growing history, \we delegates global-context modeling to the target model and performs drafting only over compact target hidden states. This design transforms speculative drafting \textbf{from an increasingly complex growing-history modeling problem into a local prediction problem} whose structure and positional range remain independent of the accumulated sequence length. \we realizes this principle through a history-free semi-autoregressive architecture and a two-stage training strategy.



\subsection{History-free semi-autoregressive Architecture} \label{sec:DLite:arch}

\begin{figure}[!t]
    \centering
    \includegraphics[width=0.98\textwidth]{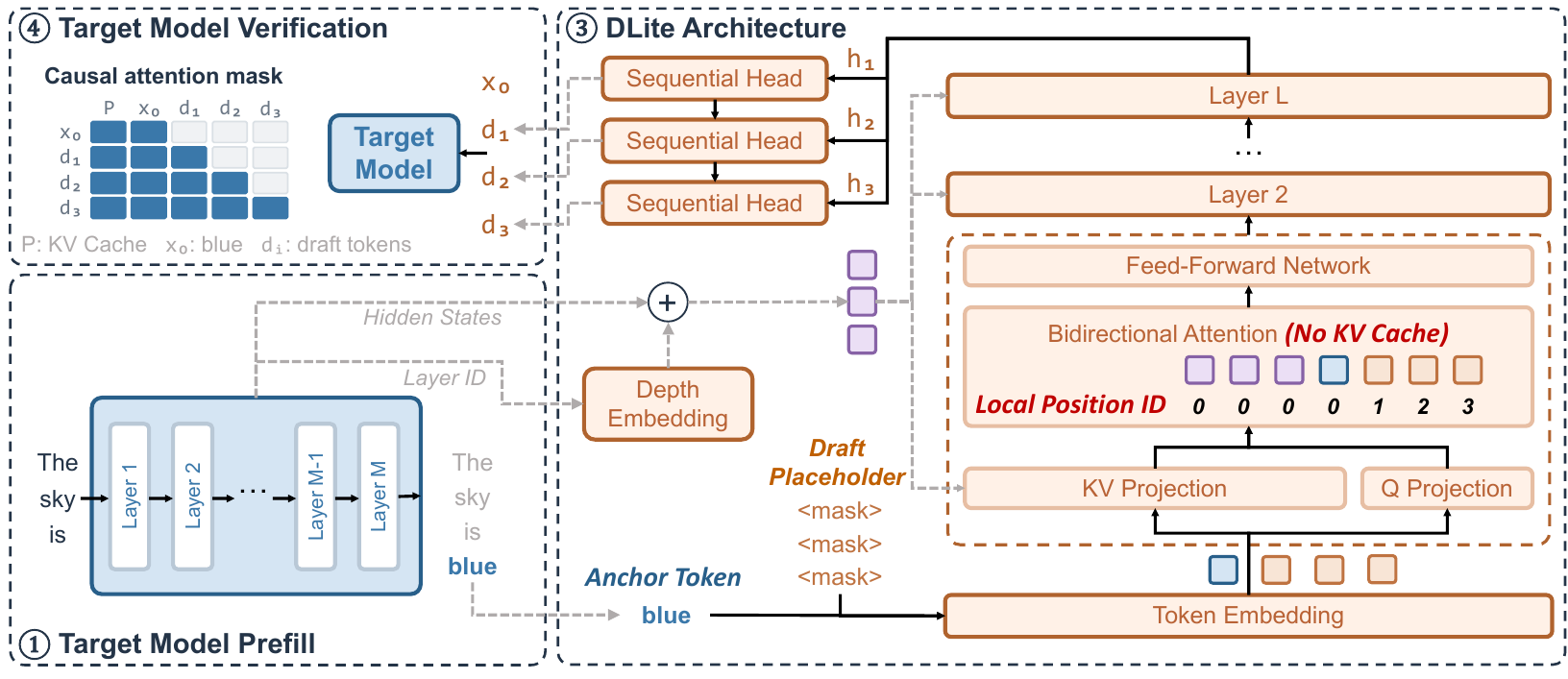}
    \vspace{-5pt}
    \caption{The inference dataflow of \we.}
    \label{fig:overview}
    \vspace{-10pt}
\end{figure}

Figure~\ref{fig:overview} illustrates the inference dataflow of \we. The target model first processes the input context and maintains its original KV cache. For each speculative round, \we takes contextual hidden states from the target model, the current anchor token, and \(K\) draft placeholders as input. A history-free parallel backbone produces hidden representations for all speculative positions simultaneously, after which a lightweight sequential head restores the causal dependency among tokens within the speculative block. The resulting candidate tokens are finally verified by the target model following the standard speculative decoding procedure. It is worth noting that \we maintains no historical draft KV cache throughout this process. Each speculative round is independently reconstructed from the latest target representations and the current anchor token, allowing the drafter to perform prediction locally without explicitly processing the accumulated history.

\subsubsection{History-free Parallel Backbone}

The key idea of the history-free parallel backbone is to \textbf{transform speculative drafting from growing-history modeling into local prediction}. To this end, the backbone combines compact target hidden states with explicit local token information and performs local prediction.


\textbf{Target hidden states.} \we directly reuses hidden states produced by the target model as compact contextual features. Since target hidden states and ordinary token embeddings correspond to different input sources, we introduce a learnable depth embedding to distinguish them. When hidden states from multiple target layers are used, the same embedding also identifies their source layers. For a target hidden state \(h_t^{(l)}\) extracted from layer \(l\), we construct
\begin{equation}
    z_t^{(l)} = h_t^{(l)} + e_{\mathrm{depth}}^{(l)}
\end{equation}
where \(e_{\mathrm{depth}}^{(l)}\) is a learnable source-specific embedding.


\textbf{Local prediction.} Conventional drafters operate over an ever-growing history, such that both the explicit context and global positions increase as decoding proceeds. \we instead delegates long-context modeling to the target model and performs draft prediction only around the current decoding position. The backbone takes the current anchor token and \(K\) speculative positions, while using local rather than globally accumulated positions. Historical information is retained through the target hidden states, allowing the drafter to focus on local prediction without explicitly reconstructing the history. \we therefore transforms growing-history modeling into a local prediction problem.

\subsubsection{Lightweight Sequential Head}

While the parallel backbone efficiently produces representations for multiple speculative positions, parallel computation removes the autoregressive dependency among tokens within the speculative block. \we therefore introduces a lightweight sequential head that \textbf{restores only the missing intra-block dependency}, rather than performing another full contextual modeling process.

Let \(t\) denote the starting position of a speculative block and \(T\) the intra-block offset. The target model predicts \(x_{t+T}\) autoregressively according to
\begin{equation}
p_T \left( x_{t+T}\mid x_{<t+T} \right) = p_T \left( x_{t+T} \mid x_{<t}, x_{t:t+T-1} \right) 
\end{equation}
The parallel backbone, however, produces \(h_T\) before the tokens at earlier speculative positions are sampled. Since the preceding context \(x_{<t}\) has already been processed by the target model and conveyed through its contextual representations into \(h_T\), the information missing from parallel drafting is precisely the intra-block prefix $ x_{t:t+T-1}$. This decomposition suggests that the sequential module does not need to reconstruct the complete historical context. It only needs to encode tokens generated earlier within the current speculative block.

To recover this missing dependency efficiently, we employ a low-rank RNN-style sequential head. It recursively compresses previously generated draft tokens into a recurrent state
\begin{equation}
s_T = g_{\phi} \left( s_{T-1}, E(x_{t+T-1}) \right)
\end{equation}
where \(E(\cdot)\) denotes the token embedding. The prediction at speculative position \(T\) is modeled as
\begin{equation}
q_{\theta,\phi} \left( x_{t+T}\mid h_T,s_T \right) \approx p_T \left( x_{t+T} \mid x_{<t}, x_{t:t+T-1} \right)
\end{equation}
Here, \(h_T\) supplies the contextual information already encoded by the target model, while \(s_T\) summarizes the missing intra-block prefix. Together, they recover the information required for autoregressive prediction without reconstructing the full history inside the draft model. This preserves most contextual computation in the parallel backbone while restricting sequential execution to a lightweight low-rank module.

Furthermore, unlike existing semi-autoregressive methods that first project every backbone representation into full-vocabulary logits and subsequently refine these predictions with an additional sequential module, \we performs sequential modeling directly in the hidden space. The vocabulary projection is applied only after the intra-block dependency has been incorporated.

For hidden dimension \(D\) and vocabulary size \(V\), this avoids repeatedly applying an intermediate \(D\times V\) projection to every speculative representation before sequential refinement. Such computation becomes particularly expensive under high-concurrency serving, where a large number of speculative hidden states are processed simultaneously.

\begin{figure}[!t]
    \centering
    \vspace{-12pt}
    \includegraphics[width=\linewidth]{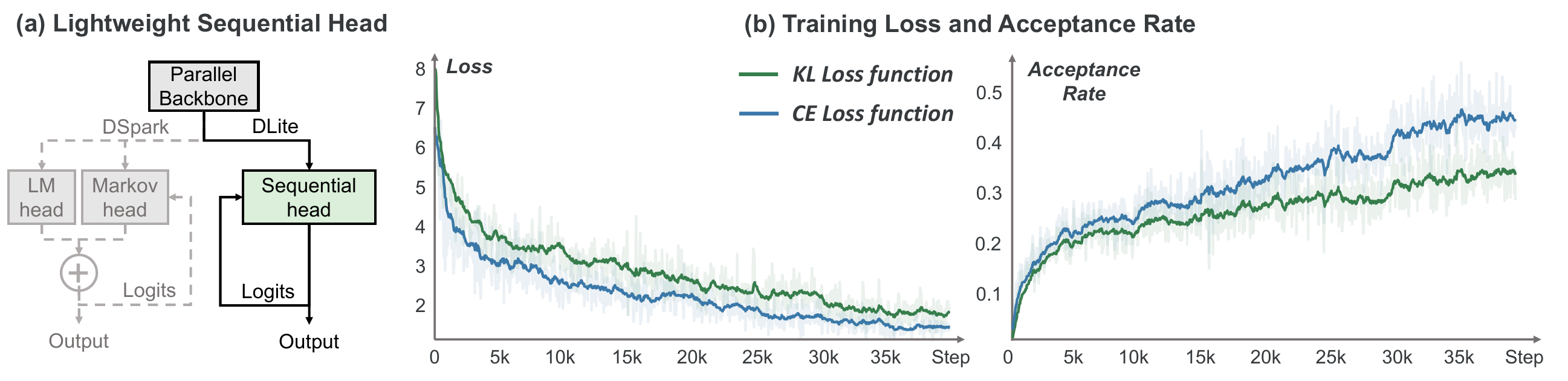}
    \caption{(a) The design architecture of the lightweight sequential head. (b) The training loss and acceptance rate curves of direct target-model distillation using KL and CE loss function.}
    \label{fig:train}
    \vspace{-15pt}
\end{figure}

\subsection{Two-stage Training Strategy} \label{sec:DLite:train}
The objective of a speculative draft model is to approximate the predictive distribution of the target model, since a larger distribution overlap generally leads to a higher speculative acceptance probability. Distribution-level objectives such as KL divergence are therefore natural choices for distillation. However, Figure~\ref{fig:train}(b) shows that directly applying KL distillation from the target model to the parallel backbone in \we performs poorly. The two models make predictions under substantially different conditions: the target model performs autoregressive prediction over the complete historical context, whereas the backbone performs parallel prediction using compressed target hidden states,
making direct distribution alignment difficult. To bridge this structural gap, we introduce a two-stage training strategy. The warm-up stage first trains \we against an isomorphic history-aware teacher, providing a structurally matched intermediate distillation target. The stable stage subsequently removes this teacher and directly aligns \we with the original target model. Let \(K\) denote the speculative block length and \(V\) the vocabulary size. At speculative position \(i\), we denote the predictive distributions of the target model, history-aware teacher, and \we as $p_i^T, p_i^H, q_i$ respectively, and denote the ground-truth token as \(y_i\).

\textbf{Warm-up Stage.} In the warm-up stage, we first train an isomorphic history-aware teacher by standard token-level cross-entropy (\(-\sum_i w_i \ \mathrm{log}\ {p_i^H(y_i)}\)). The teacher shares the same backbone and prediction structure as \we, but retains access to historical KV cache. 
Once the history-aware teacher is sufficiently trained, we use it as an intermediate distillation target for \we using forward KL divergence: 
\begin{equation}
\mathcal{L}_{\mathrm{warmup}}
=
\sum_{i=1}^{K}
w_i D_{\mathrm{KL}}
\left(
p_i^{H}\,\|\,q_i
\right)
=
\sum_{i=1}^{K}
w_i
\sum_{v\in\mathcal V}
p_i^{H}(v)
\log
\frac{p_i^{H}(v)}
{q_i(v)}
\end{equation}
where \(w_i = \exp(\frac{-(i-1)}{\gamma})\) emphasizes the contribution of earlier positions. Because the teacher and \we share the same architecture and prediction interface, their output distributions are naturally aligned despite the difference in historical information available to them.

The history-aware teacher therefore serves as an optimization bridge between the autoregressive target model and the history-free drafter. Instead of forcing \we to directly match a target distribution, the warm-up stage first exposes it to a structurally matched predictive distribution, providing a more stable starting point for subsequent target-model alignment.



\textbf{Stable Stage.} Although the intermediate teacher facilitates optimization, the ultimate objective of \we is to match the predictive distribution of the target model. We therefore discard the history-aware teacher 
in the stable stage and directly optimize \we against the target model using
\begin{equation}
\mathcal L_{\mathrm{stable}} = \lambda_{\mathrm{CE}}\mathcal L_{\mathrm{CE}} + \lambda_{\mathrm{TV}}\mathcal L_{\mathrm{TV}}
\end{equation}
where \(\mathcal L_{\mathrm{CE}}\) denotes token-level cross-entropy and
\(\mathcal L_{\mathrm{TV}}=\sum_i w_i \|p_i^T-q_i\|_1\) measures target–draft distribution discrepancy.

\section{Experiments}
\begin{table}[!t]
\centering
\caption{Acceptance length $\tau$ evaluation on general-purpose workloads.}
\label{tab:acceptance_length}
\small
\setlength{\tabcolsep}{4.2pt} 
\renewcommand{\arraystretch}{1.1}
\begin{tabular}{llccccccc} 
\toprule
& & \multicolumn{2}{c}{\textbf{Math}} & \multicolumn{3}{c}{\textbf{Code}} & \multicolumn{2}{c}{\textbf{Chat}} \\ 
\cmidrule(lr){3-4} \cmidrule(lr){5-7} \cmidrule(lr){8-9}          
\textbf{Model} & \textbf{Method} & \textbf{GSM8K} & \textbf{MATH-500} & \textbf{HumanEval} & \textbf{\ MBPP} & \textbf{\ \ \ LCB \ \ } & \textbf{MT-Bench} & \textbf{Alpaca} \\
\midrule
\multirow{6}{*}{Qwen3-4B}   & EAGLE3 & 2.63 & 2.39 & 2.47 & 2.34 & 2.17 & 3.07 & 2.28 \\
                            & DFlow  & 5.13 & 4.92 & 4.33 & 4.12 & - & 3.80 & 2.90 \\
                            & DFlash    & 5.97 & 5.82 & 5.25 & 4.97 & 4.82 & 3.39 & 3.56 \\
                            & DFlash2    & 6.14 & 5.80 & 5.38 & 5.14 & 4.79 & \textbf{4.14} & 2.50 \\
                            & DSpark    & 6.27 & 6.25 & 5.62 & 5.48 & 5.20  & 3.81 & 3.54 \\
                            & DLite     & \textbf{6.38} & \textbf{6.30} & \textbf{5.68} & \textbf{5.64} & \textbf{5.31} & 3.94 & \textbf{3.64} \\
\midrule
\multirow{5}{*}{Qwen3-8B}   & EAGLE3 & 2.60 & 2.28 & 2.51 & 2.17 & 2.08 & 3.07 & 2.56 \\
                            & DFlow  & 5.10 & 4.94 & 4.28 & 4.08 & - & 3.58 & 2.93 \\
                            & DFlash    & 6.04 & 5.91 & 5.27 & 4.88 & 4.89 & 3.39 & 3.56 \\
                            & DSpark    & 6.36 & 6.34 & 5.86 & 5.36 & 5.43 & 3.88 & 3.64 \\
                            & DLite     & \textbf{6.48} & \textbf{6.35} & \textbf{5.96} & \textbf{5.65} & \textbf{5.66} & \textbf{3.96} & \textbf{3.78} \\
\bottomrule
\end{tabular}
\end{table}

\subsection{Experimental Setup} \label{sec:exp:setup}


\textbf{Models and Baselines.} We evaluate \we on Qwen3-4B and Qwen3-8B. We compare against vanilla autoregressive decoding (baseline) and representative speculative decoding methods, including EAGLE-3~\cite{li2026eagle3}, DFlash~\cite{chen2026dflash}, DFlash2~\cite{dflash2}, DSpark~\cite{cheng2026dspark}, and DFlow~\cite{zhang2026dflow}. All serving experiments are conducted on NVIDIA H800 GPUs. The results of DFlash2 and DFlow are from RedHatAI~\cite{redhat} or paper.


\textbf{Benchmarks.} We evaluate the performance in three different domains, general-purpose tasks, long-context tasks and agentic tasks. General-purpose tasks include three categories, Math: GSM8K~\cite{gsm8k} and MATH-500~\cite{math500}, Code: HumanEval~\cite{humaneval}, MBPP~\cite{MBPP}, and LiveCodeBench~\cite{jain2025livecodebench}, Chat: MT-Bench~\cite{zheng2023judging} and Alpaca~\cite{alpaca}. Long-context tasks include Multi-Document QA, Long In-context Learning, Code Repository Understanding, and Long dialogue History in LongBench\_v2~\cite{bai2024longbench}. Agentic tasks include AgentMemoryBench~\cite{liu2024agentbench}, AgentLongBench~\cite{fang2026agentlongbenchcontrollablelongbenchmark}, BFCL-v3~\cite{bfcl} and SWE-Bench~\cite{jimenez2024swebench}.

\textbf{Datasets.}
We use Open-PerfectBlend~\cite{xu2024perfect} which consists of 1.3 million samples containing chat, math, code and instruction-following data. We also collect 600K samples from NVIDIA Nemotron Post-Training Dataset V2~\cite{NemotronPostTrainingDatasetV2} to augment the dataset. We only use the prompts from the dataset and generate the response using the target models.

\textbf{Implementation Details.} For all speculative decoding methods, we use a speculative block size of \(K=7\) and follow the original implementation whenever available. \we uses 12/14 layer target hidden states with sequence-wise concatenation, and a sequential-head rank of \(R=320/512\) for Qwen3-4B/8B. We report average acceptance length \(\tau\) and end-to-end decoding speedup over vanilla autoregressive decoding. 

\subsection{Performance Evaluation}

\textbf{General-purpose workloads.} Table~\ref{tab:acceptance_length}  reports acceptance lengths on conventional reasoning, coding, and conversational benchmarks. \we consistently matches or outperforms existing speculative drafters on both Qwen3-4B and Qwen3-8B. These results show that removing the historical draft KV cache does not sacrifice drafting quality on standard workloads, despite \we operating only on target representations and a local anchor token.


\textbf{Long-context workloads.} 
As shown in Table~\ref{tab:longbench_v2_models}, the advantage of \we becomes more pronounced on LongBench v2, where it consistently improves both acceptance length and end-to-end speedup over DFlash and DSpark. This trend supports our central design motivation: as the context grows, conventional drafters increasingly rely on their own long-history modeling, whereas \we delegates global-context modeling to the target model and performs only local prediction at the draft side.

\textbf{Agentic workloads.} 
Table~\ref{tab:agent_benchmarks} shows the largest relative gains under agentic workloads. In particular, on SWE-Bench, \we substantially improves acceptance length over DFlash and DSpark while achieving around \(2\times\) end-to-end speedup. These results indicate that history-free drafting is especially beneficial under the long, evolving contexts characteristic of agent execution

\begin{table}[!t]
\centering
\vspace{-10pt}
\caption{Speedup and acceptance length ($\tau$) evaluation on long-context workloads.}
\label{tab:longbench_v2_models}
\small
\renewcommand{\arraystretch}{1.1}
\setlength{\tabcolsep}{5.8pt} 
\begin{tabular}{c l cc cc cc cc}
\toprule
\multirow{3}{*}{\textbf{Model}} & \multirow{3}{*}{\textbf{Method}} & \multicolumn{8}{c}{\textbf{LongBench\_v2}} \\
\cmidrule(lr){3-10}
& & \multicolumn{2}{c}{\textbf{Multi-Doc QA}} & \multicolumn{2}{c}{\textbf{Long In-ctx Lrn.}} & \multicolumn{2}{c}{\textbf{Code Repo Ud.}} & \multicolumn{2}{c}{\textbf{Long-dialog Hist.}} \\
\cmidrule(lr){3-4} \cmidrule(lr){5-6} \cmidrule(lr){7-8} \cmidrule(lr){9-10}
& & Speedup & $\tau$ & Speedup & $\tau$ & Speedup & $\tau$ & Speedup & $\tau$ \\
\midrule

\multirow{4}{*}{Qwen3-4B} 
& Eagle3      & 1.70$\times$ & 2.10  & 1.76$\times$  & 2.28   & 1.68$\times$  & 2.12 & 2.23$\times$  & 2.88   \\
& DFlash     & 1.60$\times$  & 1.90 & 1.71$\times$ & 2.00 & 1.45$\times$ & 1.83 & 2.62$\times$ & 3.04 \\
& DSpark     & 1.81$\times$ & 2.20 & 2.01$\times$ & 2.41 & 2.02$\times$ & 2.44 & 3.16$\times$ & 3.93 \\
& DLite & \textbf{2.18$\times$} & \textbf{2.74} & \textbf{2.58$\times$} & \textbf{3.11} & \textbf{2.66$\times$} & \textbf{3.14} & \textbf{3.88$\times$} & \textbf{4.62} \\

\midrule

\multirow{4}{*}{Qwen3-8B}
& Eagle3      & 1.64$\times$ & 2.05 & 1.87$\times$ & 2.33 & 1.42$\times$  & 1.80 & 2.01$\times$  & 2.56 \\
& DFlash     & 1.81$\times$  & 2.10 & 1.86$\times$ & 2.28 & 1.65$\times$  & 2.16 & 2.58$\times$  & 3.07 \\
& DSpark     & 1.98$\times$  & 2.52 & 2.28$\times$ & 2.67 & 2.02$\times$  & 2.59 & 3.02$\times$  & 4.01 \\
& DLite & \textbf{2.32$\times$} & \textbf{2.86} & \textbf{2.74$\times$} & \textbf{3.22} & \textbf{2.53$\times$} & \textbf{3.04} & \textbf{3.96$\times$} & \textbf{4.77} \\ 
\bottomrule
\end{tabular}
\end{table}

\begin{table}[!h]
\centering
\vspace{-10pt}
\caption{Speedup and acceptance length ($\tau$) evaluation on agentic workloads.}
\label{tab:agent_benchmarks}
\small
\renewcommand{\arraystretch}{1.1}
\setlength{\tabcolsep}{4pt}
\begin{tabular}{l l cc cc cc cc}
\toprule
\multirow{2}{*}{\textbf{Metric}}
& \multirow{2}{*}{\textbf{Method}}
& \multicolumn{2}{c}{\textbf{AgentMemBench}}
& \multicolumn{2}{c}{\textbf{AgentLongBench}}
& \multicolumn{2}{c}{\textbf{\ \ \ \ \ \ \ \ \ BFCL-v3 \ \ \ \ \ \ \ \ \ }}
& \multicolumn{2}{c}{\textbf{\ \ \ \ SWE-bench\ \ \ \ }} \\
\cmidrule(lr){3-4}
\cmidrule(lr){5-6}
\cmidrule(lr){7-8}
\cmidrule(lr){9-10}
& & \ \ Speedup & $\tau$
& Speedup & $\tau$
& Speedup & $\tau$
& Speedup & $\tau$ \\
\midrule

\multirow{4}{*}{Qwen3-4B} 
& Eagle3      & 1.66$\times$   & 2.09   & 1.70$\times$  & 2.12  & 1.58$\times$  & 2.00   & 1.48$\times$  & 1.88   \\
& DFlash     & 1.22$\times$  & 1.65 & 0.95$\times$ & 1.14 & 1.44$\times$ & 1.80 & 1.47$\times$ & 1.72 \\
& DSpark     & 1.72$\times$ & 2.10 & 1.45$\times$ & 1.84 & 2.11$\times$ & 2.54 & 1.64$\times$ & 1.89 \\
& DLite & \textbf{2.39$\times$} & \textbf{2.80} & \textbf{2.07$\times$} & \textbf{2.51} & \textbf{2.37$\times$} & \textbf{2.78} & \textbf{2.68$\times$} & \textbf{3.15} \\

\midrule

\multirow{4}{*}{Qwen3-8B}
& Eagle3      &1.77$\times$ & 2.19 & 1.60$\times$ & 2.00 & 1.49$\times$  & 1.88 & 1.33$\times$  & 1.71 \\
& DFlash     & 1.80$\times$  & 2.20 & 1.28$\times$ & 1.65 & 1.78$\times$  & 2.07 & 1.53$\times$  & 1.83 \\
& DSpark     & 1.83$\times$  & 2.26 & 1.68$\times$ & 2.09 & 1.82$\times$  & 2.23 & 1.60$\times$  & 1.94 \\
& DLite & \textbf{2.20$\times$} & \textbf{2.69} & \textbf{2.31$\times$} & \textbf{2.79} & \textbf{2.30$\times$} & \textbf{2.76} & \textbf{2.74$\times$} & \textbf{3.19} \\

\bottomrule
\end{tabular}
\end{table}




\textbf{High-concurrency throughput evaluation.} Table~\ref{tab:throughput} evaluates end-to-end throughput under concurrency from 1 to 64. \we consistently achieves higher throughput than DSpark across both model sizes and all evaluated workloads. Notably, the throughput gain remains visible even when their average acceptance lengths are close, showing that the improvement does not solely come from better draft quality. This result is consistent with the lower drafting overhead of our lightweight sequential head, which avoids the intermediate full-vocabulary projection.

\begin{table*}[t]
\centering
\vspace{-10pt}
\caption{
High-concurrency throughput evaluation on Qwen3 models.
}
\label{tab:throughput}
\renewcommand{\arraystretch}{1.1}
\setlength{\tabcolsep}{4pt}
\resizebox{\textwidth}{!}{
\begin{tabular}{llccccc|c}
\toprule
\multirow{2}{*}{\textbf{Task}} &
\multirow{2}{*}{\textbf{Method}}
& \multicolumn{5}{c|}{\textbf{Concurrency (TPS / Speedup)}}
& \multirow{2}{*}{\textbf{Avg. $\tau$}} \\
\cmidrule(lr){3-7}
& & 1 & 8 & 16 & 32 & 64 & \\
\midrule

\multicolumn{8}{l}{\textbf{Qwen3-4B}} \\
\midrule

\multirow{3}{*}{Alpaca}
& Baseline & 246 / 1.00$\times$ & 1798 / 1.00$\times$ & 3377 / 1.00$\times$ & 5849 / 1.00$\times$ & 10808 /1.00$\times$ & -- \\

& DSpark & 609 / 2.47$\times$ & 4131 / 2.30$\times$ & 7018 / 2.08$\times$ & 11118 / 1.90$\times$ & 13548 / 1.25$\times$ & 3.49 \\

& \textbf{DLite} & \textbf{635 / 2.58$\times$} & \textbf{4222 / 2.35$\times$} & \textbf{7293 / 2.16$\times$} & \textbf{11792 / 2.02$\times$} & \textbf{14300 / 1.32$\times$} & \textbf{3.63} \\

\midrule

\multirow{3}{*}{MT-Bench}
& Baseline
& 245 / 1.00$\times$
& 1792 / 1.00$\times$
& 3377 / 1.00$\times$
& 5883 / 1.00$\times$
& 10113 / 1.00$\times$
& -- \\

& DSpark
& 620 / 2.53$\times$
& 4589 / 2.56$\times$
& 8177 / 2.42$\times$
& 12565 / 2.14$\times$
& 15830 / 1.57$\times$
& 3.79 \\

& \textbf{DLite}
& \textbf{633 / 2.58$\times$}
& \textbf{4699 / 2.62$\times$}
& \textbf{8162 / 2.42$\times$}
& \textbf{12994 / 2.21$\times$}
& \textbf{16820 / 1.66$\times$}
& \textbf{3.88} \\

\midrule

\multirow{3}{*}{LCB}
& Baseline & 242 / 1.00$\times$ & 1665 / 1.00$\times$ & 2918 / 1.00$\times$ & 4498 / 1.00$\times$& 6586 / 1.00$\times$ & -- \\

& DSpark & 905 / 3.74$\times$ & 5457 / 3.28$\times$ & 8248 / 2.83$\times$ & 10243 / 2.28$\times$ & 11309 / 1.72$\times$ & 5.28 \\

& \textbf{DLite} & \textbf{960 / 3.97$\times$} & \textbf{5592 / 3.36$\times$} & \textbf{8418 / 2.88$\times$} & \textbf{10873 / 2.42$\times$} & \textbf{12075 / 1.83$\times$} & \textbf{5.98} \\

\midrule

\multicolumn{8}{l}{\textbf{Qwen3-8B}} \\
\midrule

\multirow{3}{*}{Alpaca}
& Baseline
& 142 / 1.00$\times$
& 993 / 1.00$\times$
& 1705 / 1.00$\times$
& 2920 / 1.00$\times$
& 4330 / 1.00$\times$
& -- \\

& DSpark
& 364 / 2.56$\times$
& 2260 / 2.28$\times$
& 3758 / 2.20$\times$
& 5238 / 1.79$\times$
& 6827 / 1.58$\times$
& 3.69 \\

& \textbf{DLite}
& \textbf{379 / 2.67$\times$}
& \textbf{2409 / 2.43$\times$}
& \textbf{3965 / 2.33$\times$}
& \textbf{5325 / 1.82$\times$}
& \textbf{7042 / 1.63$\times$}
& \textbf{3.86} \\

\midrule

\multirow{3}{*}{MT-Bench}
& Baseline
& 143 / 1.00$\times$
& 977 / 1.00$\times$
& 1749 / 1.00$\times$
& 2950 / 1.00$\times$
& 4450 / 1.00$\times$
& -- \\

& DSpark
& 398 / 2.78$\times$
& 2454 / 2.51$\times$
& 3956 / 2.26$\times$
& 5455 / 1.85$\times$
& 7236 / 1.63$\times$
& 3.89 \\

& \textbf{DLite}
& \textbf{407 / 2.85$\times$}
& \textbf{2531 / 2.59$\times$}
& \textbf{4131 / 2.36$\times$}
& \textbf{5558 / 1.88$\times$}
& \textbf{7305 / 1.64$\times$}
& \textbf{4.10} \\

\midrule

\multirow{3}{*}{LCB}
& Baseline
& 142 / 1.00$\times$
& 902 / 1.00$\times$
& 1401 / 1.00$\times$
& 2196 / 1.00$\times$
& 2708 / 1.00$\times$
& -- \\

& DSpark
& 511 / 3.60$\times$
& 2562 / 2.84$\times$
& 3803 / 2.71$\times$
& 4815 / 2.19$\times$
& 5621 / 2.08$\times$
& 5.47 \\

& \textbf{DLite}
& \textbf{561 / 3.95$\times$}
& \textbf{2761 / 3.06$\times$}
& \textbf{4016 / 2.87$\times$}
& \textbf{5059 / 2.30$\times$}
& \textbf{5758 / 2.13$\times$}
& \textbf{6.04} \\

\bottomrule
\end{tabular}
}
\end{table*}

\subsection{Ablation Study and Design Space Exploration}

\begin{figure}[!h]
    \centering
    \includegraphics[width=\linewidth]{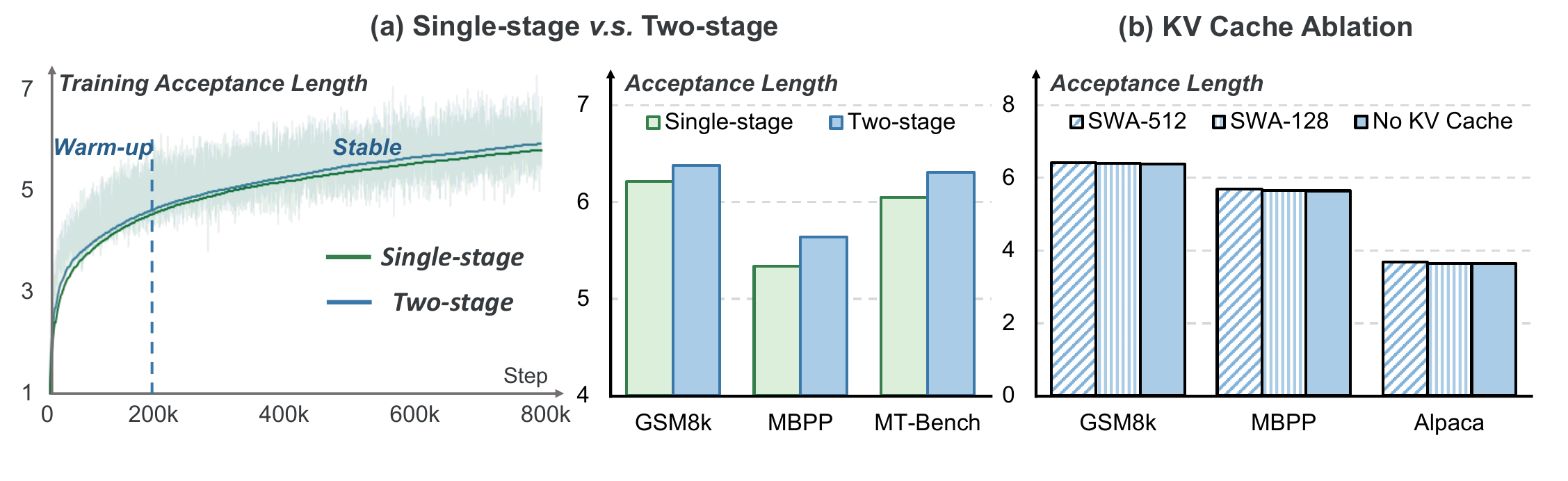}
    \vspace{-20pt}
    \caption{Ablation on training strategy and KV cache.}
    \label{fig:ablation}
\end{figure}

\textbf{Training strategy.}  Figure~\ref{fig:ablation}(a) compares our two-stage training strategy with direct single-stage training. The two-stage strategy consistently achieves higher acceptance length throughout the stable stage and improves the final performance. This result supports our motivation in Section~\ref{sec:DLite:train}: the history-aware isomorphic teacher provides an effective optimization bridge before direct alignment.

\textbf{Historical KV cache.} To examine whether historical draft states are necessary for accurate speculation, we additionally train two variants with sliding-window KV caches of 128 and 512 tokens using the same training data and setup as \we. As shown in Figure~\ref{fig:ablation}(b), after sufficient training, the three models achieve comparable acceptance lengths across the evaluated benchmarks. This suggests that historical KV states can facilitate prediction at earlier stages of training, but their advantage diminishes as the drafter learns to exploit contextual information transferred through the target hidden states. With sufficient training, \we can therefore achieve comparable quality without explicitly maintaining KV cache.

\textbf{Design space exploration.} We further explore several architectural design choices of \we, including the number of hidden states extracted from the target model, how multiple target hidden states are composed (feature-wise versus sequence-wise concatenation), and the rank of the lightweight sequential head. These dimensions jointly determine the trade-off between contextual representation capacity, drafting quality, and execution overhead. We use this design-space exploration in Appendix~\ref{app:dse} to determine the final configurations reported in Section~\ref{sec:exp:setup}.

\section{Conclusion}

In this work, we present \we, a history-free speculative decoding framework for agentic AI. \we removes the need for maintaining historical draft KV cache by directly reusing contextual hidden states from the target model and reformulating draft generation as local speculative prediction. On top of the history-free parallel backbone, a lightweight sequential head restores intra-block causal dependencies with low overhead, while a tailored two-stage training strategy addresses the optimization difficulty caused by the heterogeneous prediction structures of the draft and target models. Extensive experiments on general-purpose, long-context, and agentic workloads show that \we achieves \(2\sim6\) acceptance length and improves over the state-of-the-art semi-autoregressive drafter DSpark by up to \(70\%\), while eliminating historical draft KV-cache maintenance. These results demonstrate that removing redundant historical modeling from the drafter can simultaneously improve speculative decoding quality and system efficiency for emerging agentic workloads.

\subsubsection*{Acknowledgments}
This work was sponsored by the National Natural Science Foundation of China (No. 62622406, 62674144, 62561160156) and Shanghai Rising-Star Program (No. 24QB2706200).

\bibliography{iclr2027_conference}
\bibliographystyle{iclr2027_conference}

\appendix
\section{Appendix}
\subsection{Training Details}

We train \we on the datasets described in Section~\ref{sec:exp:setup}. For all training data, we only use the original prompts and regenerate responses using the corresponding target model, ensuring that the supervision is consistent with the target distribution.

\textbf{Model configuration.} For Qwen3-4B and Qwen3-8B, the \we backbone contains 5 Transformer layers with hidden dimension \(2560/4096\). We extract 12/14 hidden states from the target model using sequence-wise composition. Specifically, we always retain the hidden states from the first and last layers, while uniformly sampling the remaining hidden states from the intermediate layers. The rank of lightweight sequential head is \(R=320/512\) for Qwen3-4B/8B, respectively. The speculative block length is set to \(K=7\).

\textbf{Training data.} Our training corpus contains approximately 1.9M samples in total, including 1.3M samples from Open-PerfectBlend and 600K samples from NVIDIA Nemotron Post-Training Dataset V2. We use only the prompts from these datasets and generate responses with the target model. The maximum sequence length is 4096, and each training sample contains 512 speculative blocks / anchors on average.

\textbf{Optimization.} We train \we using AdamW with a learning rate of $4 \times 10^{-4}$ and gradient clipping threshold 1.0. The global batch size is 32, with gradient accumulation of 1 step. We use cosine learning-rate decay with 0.04 warm-up ratio. Training is performed on 32 NVIDIA H800 GPUs.

\textbf{Two-stage training.} The positional weight decay $\gamma$ is set to 4. Training consists of a warm-up stage followed by a stable stage. In the warm-up stage, we first train the isomorphic history-aware teacher with the same data mentioned above using token-level cross-entropy for 8 epochs. We then train \we against this teacher using forward KL divergence for 2 epochs. In the stable stage, the history-aware teacher is removed and \we is directly optimized against the target model using
\[
\mathcal{L}_{\mathrm{stable}}
=
\lambda_{\mathrm{CE}}\mathcal{L}_{\mathrm{CE}}
+
\lambda_{\mathrm{TV}}\mathcal{L}_{\mathrm{TV}},
\]
where \(\lambda_{\mathrm{CE}}=0.1\) and \(\lambda_{\mathrm{TV}}=1\). The stable stage lasts for 6 epochs.

\subsection{Design Space exploration} \label{app:dse}
We conduct a design space exploration on Qwen3-8B to study three key architectural choices of \we: (1) the number of target hidden states used as contextual features, (2) the composition strategy for multiple target hidden states, and (3) the rank of the lightweight sequential head. To reduce the cost of design exploration, all variants in this section are trained on the same 80K-sample subset, while keeping the remaining training configurations unchanged. We evaluate the average acceptance length on Alpaca, GSM8K, MBPP, and Multi-Document QA, covering conversational, mathematical, coding, and long-context workloads.

\begin{table*}[h]
\centering
\caption{
Design space exploration of \we on Qwen3-8B.
All variants are trained on the same 80K-sample subset, and we report average acceptance length ($\tau$).
For the target hidden-state exploration, the first and last target layers are always included in addition to the sampled intermediate layers.
Unless otherwise specified, we use 14 target hidden states, sequence-wise hidden-state composition, and a sequential-head rank of 512.
Bold indicates the configuration adopted in the final model.
}
\label{tab:dse}
\resizebox{\textwidth}{!}{
\begin{tabular}{llccccc}
\toprule
\textbf{Dimension} &
\textbf{Configuration} &
\textbf{Alpaca} &
\textbf{GSM8K} &
\textbf{MBPP} &
\textbf{Multi-Doc QA} &
\textbf{Avg. $\tau$} \\
\midrule

\multirow{5}{*}{\textbf{Target Hidden States}}
& 7
& 2.34 & 4.36 & 3.63 & 2.22 & 3.14 \\

& 9
& 2.42 & 4.40 & 3.75 & 2.24 & 3.20 \\

& \textbf{14}
& \textbf{2.63} & \textbf{4.53} & \textbf{3.76} & \textbf{2.28} & \textbf{3.30} \\

& 20
& 2.62 & 4.55 & 3.74 & 2.26 & 3.29 \\

& 36
& 2.58 & 4.51 & 3.71 & 2.26 & 3.27 \\

\midrule

\multirow{2}{*}{\textbf{Hidden Composition}}
& Feature-wise
& 2.20 & 4.28 & 3.50 & 2.16 & 3.04 \\

& \textbf{Sequence-wise}
& \textbf{2.63} & \textbf{4.53} & \textbf{3.76} & \textbf{2.28} & \textbf{3.30} \\

\midrule

\multirow{5}{*}{\textbf{Sequential-Head Rank}}
& 256 ($1/16D$)
& 2.48 & 4.40 & 3.61 & 2.16 & 3.16 \\

& \textbf{512 ($1/8D$)}
& \textbf{2.63} & \textbf{4.53} & \textbf{3.76} & \textbf{2.28} & \textbf{3.30} \\

& 1024 ($1/4D$)
& 2.64 & 4.55 & 3.80 & 2.28 & 3.32 \\

& 2048 ($1/2D$)
& 2.64 & 4.53 & 3.82 & 2.28 & 3.32 \\

& 4096 ($D$)
& 2.64 & 4.56 & 3.82 & 2.30 & 3.33 \\

\bottomrule
\end{tabular}
}
\end{table*}

\textbf{Number of target hidden states.} Table~\ref{tab:dse} shows that increasing the number of target hidden states initially leads to clear improvements in acceptance length. The gains, however, quickly saturate: increasing the number from 7 to 14 improves performance consistently across all four workloads, while using 20 or 36 hidden states provides little or no additional benefit. This suggests that a moderate number of intermediate target representations is sufficient to capture complementary contextual information, while further increasing the number of features introduces redundant representations and additional computation.

\textbf{Hidden-state composition.} We further compare two ways of combining hidden states extracted from different target layers. Sequence-wise concatenation treats representations from different layers as separate input elements, while feature-wise concatenation concatenates them along the hidden dimension. Sequence-wise concatenation consistently outperforms feature-wise concatenation across all evaluated workloads, increasing the average acceptance length from 3.04 to 3.30. We therefore adopt sequence-wise composition in \we. This result also supports the use of depth embeddings in Section~\ref{sec:DLite:arch}, which explicitly identify the source layers of hidden states when they are presented as separate inputs.

\textbf{Sequential-head rank.} Finally, we vary the rank of the lightweight sequential head from 1/16 of the hidden dimension (256) to the full hidden dimension (4096). Increasing the rank from 256 to 512 yields a noticeable improvement in drafting quality while increasing the end-to-end latency by only 0.3\%, whereas larger ranks provide only marginal gains while introducing substantial computational overhead. This suggests that the information required by the draft model can be sufficiently resolved in a low-rank space. For example, the average acceptance length increases from 3.16 at rank 256 to 3.30 at rank 512, but only reaches 3.33 with the full 4096-dimensional head. Considering the additional sequential computation introduced by a larger rank and the diminishing performance gains, we choose a relatively small rank to balance drafting quality and execution efficiency.

\paragraph{Implementation details.}
We evaluate \we on four agent-oriented benchmarks using Qwen3-4B and Qwen3-8B targets. For SWE-Bench, we use 50 prompts from 64K settings (40K--64K input tokens). For AgentLongBench, we use 50 prompts from the 64K setting (55.5K--74.1K input tokens). For MemoryAgentBench, we select 100 FactConsolidation, 50 EventQA, and 30 free-form DetectiveQA questions (66K--118K input tokens), constructing one independent long-context prompt per question. For BFCL, we use 41 sessions from the \texttt{multi\_turn\_long\_context} category, run its multi-turn tool-execution harness through a local OpenAI-compatible inference adapter, and aggregate model calls whose input exceeds 32K tokens. BFCL tool-execution time is excluded.

\end{document}